\documentclass{article}
\usepackage[dvipsnames]{xcolor}

\usepackage[utf8]{inputenc}
\usepackage{natbib}

\usepackage[hyperref]{emnlp2021}

\usepackage{times}
\usepackage{latexsym}
\usepackage{graphicx}
\usepackage{booktabs}
\usepackage{subfig}
\usepackage[textsize=large]{todonotes}
\usepackage{booktabs}
\usepackage{adjustbox}
\usepackage{algorithm}
\usepackage{algpseudocode}
\usepackage[inline]{enumitem}
\usepackage{amsmath,amssymb}
\usepackage[font=small,skip=5pt]{caption}
\usepackage{tabularx} 
\usepackage{longtable}
\usepackage{url}
\usepackage{cleveref}
\usepackage{enumitem}
\usepackage{listings}
\usepackage[compact]{titlesec}
\usepackage{multirow}
\usepackage{wasysym}
\usepackage{wrapfig}
\usepackage{xspace}
\usepackage{listings}

\newcommand{\removed}[1]{}
\newcommand{\REMOVED}[1]{}

\newcommand{\poincare}{Poincar\'{e}\xspace}

\title{Comparing Euclidean and Hyperbolic Embeddings on the WordNet Nouns Hypernymy Graph}

\author{
  Sameer Bansal, Adrian Benton \\
  Bloomberg \\
  731 Lexington Ave \\
  New York, NY 10022 USA \\
  {\tt \{sbansal70,abenton10\}@bloomberg.net} \\
}
\date{\today}

\begin{document}

\maketitle

\begin{abstract}
\citet{nickel2017poincare} present a new method for embedding tree nodes in the \poincare ball, and suggest that these hyperbolic embeddings are far more effective than Euclidean embeddings at embedding nodes in large, hierarchically structured graphs like the WordNet nouns hypernymy tree. 
This is especially true in low dimensions
~\citep[Table~1]{nickel2017poincare}.
In this work, we seek to reproduce their experiments on embedding and reconstructing the WordNet nouns hypernymy graph. 
Counter to what they report, we find that Euclidean embeddings are able to represent this tree at least as well as \poincare embeddings, when allowed at least 50 dimensions. 
We note that this does not diminish the significance of their work given the impressive performance of hyperbolic embeddings in very low-dimensional settings.
However, given the wide influence of their work, our aim here is to present an updated and more accurate comparison between the Euclidean and hyperbolic embeddings.


\end{abstract}

\section{Introduction}
\label{sec:introduction}

\citet{nickel2017poincare} introduced a method for learning embeddings in hyperbolic space for large, hierarchically structured objects like the WordNet nouns hypernymy graph.  
This work convincingly shows that across a range of embedding dimensions, from as low as 5 to as high as 200, hyperbolic embeddings consistently outperformed their Euclidean counterparts~\citep[Table~1]{nickel2017poincare}.
Illustrating the difference in performance at the highest experimental setting of 200 dimensions, the mean average precision (MAP) score for hyperbolic embeddings was shown to be around 5 times that of Euclidean for embedding nouns in the WordNet hypernymy graph.\footnote{The embeddings were evaluated on a reconstruction task where a MAP score closer to 1 indicates better performance.}
These experiments have been extremely influential, with the results on embedding the WordNet nouns hypernymy graph baselines often cited in later works on enhanced hyperbolic embeddings~\citep{de2018representation,ganea2018hyperbolic,dhingra-etal-2018-embedding,lopez-etal-2019-fine,balazevic2019multi,feyisetan2019leveraging,chami2020low}.\footnote{\citet{nickel2017poincare} has been cited over 500 times (source: {\tt www.semanticscholar.org}).}

In this work, we reproduce the reconstruction error experiments on the WordNet noun hypernymy graph from~\citet{nickel2017poincare}.
Counter to what they report, we find that Euclidean word embeddings are as effective at encoding the WordNet nouns graph as hyperbolic embeddings when given at least 50 dimensions. 
In fact, Euclidean embeddings with $\geq 100$ dimensions achieve lower reconstruction error over embeddings in the Lorentz model, an improved hyperbolic embedding method, which was published the following year \citep{nickel2018learning}.

The inability to reproduce the reported Euclidean experiments has been raised in several issues in the associated GitHub repository.\footnote{\url{https://github.com/facebookresearch/poincare-embeddings/issues/35} ; 68 ; 72}
This has also been acknowledged by the authors of the original study, who suggest that the original Euclidean embeddings were regularized in a way that may have hurt performance.\footnote{Author response on GitHub \url{github.com/facebookresearch/poincare-embeddings/issues/35\#issuecomment-685174866}}
However, the published manuscript has not been updated to reflect these problems with reproducing the Euclidean embedding baselines. As such, we hope that our reproduction will serve as a useful reference for those who are interested in exploring hyperbolic embeddings.

\section{Experimental Setup}
\label{sec:experiments}


We use the source code released by the authors to carry out all experiments in this study, reusing the data processing, model training, and evaluation pipelines.

\vspace{.1in}
\noindent {\bf Dataset.} Following~\citet{nickel2017poincare}, we embed the WordNet noun hierarchy~\citep[{\em WordNet-Nouns}]{wordnetonline} in both Euclidean and hyperbolic space.
Though the original study also published results on additional datasets, we restrict our focus to {\em WordNet-Nouns}, which exhibited a considerable gap in performance between embeddings.

\begin{figure}[t]
  \centering
  \includegraphics[width=0.9\linewidth]{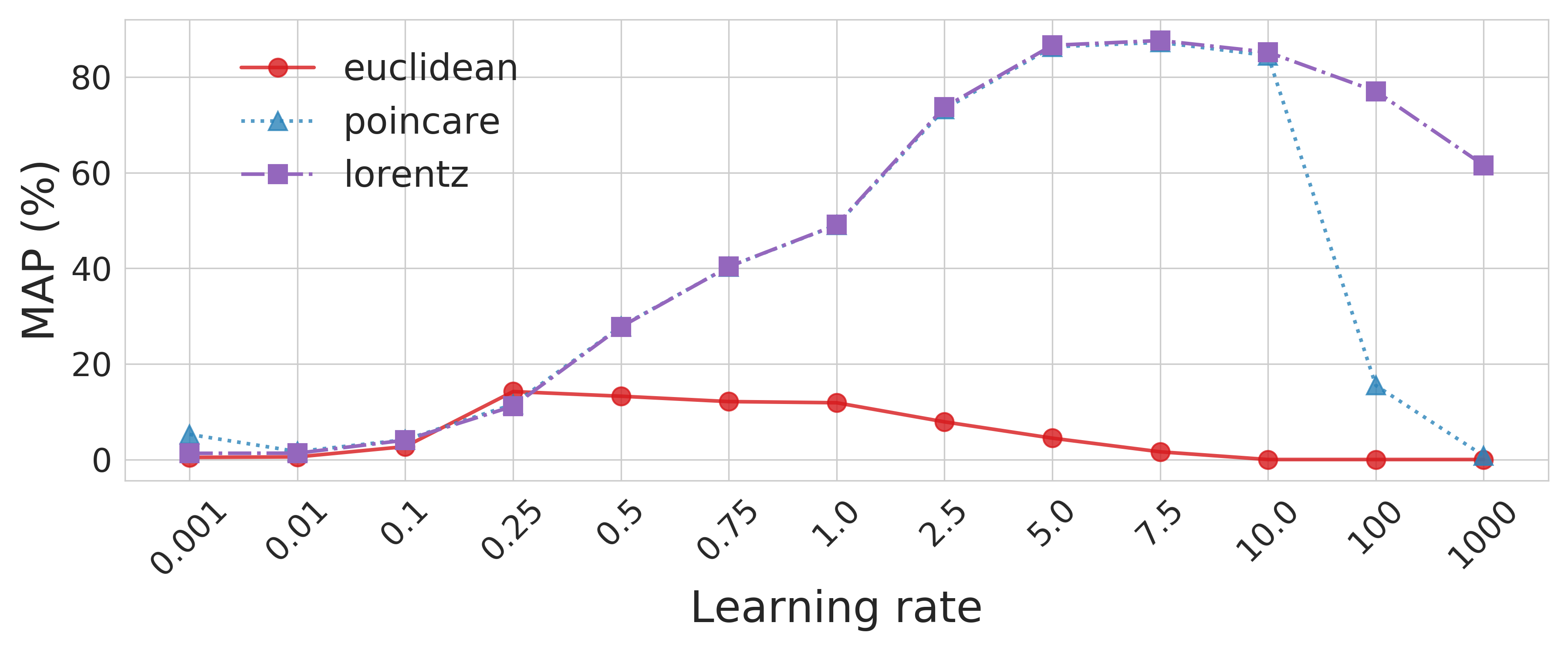}
  \caption{MAP of Euclidean, \poincare, and Lorentz 20-dimensional embeddings as a function of learning rate. MAP scores for \poincare and Lorentz embeddings are very similar up to a learning rate of 10.0.}
  \label{fig:lr_sweep}
\end{figure}

\vspace{.1in}
\noindent {\bf Model training.} 
Other than learning rate, we retain the default hyperparameters specified in the released source code, and train embeddings for 1,500 epochs.
The learning rate is tuned in the range $10^{[-2, 3]}$
independently for each class of embeddings (Figure~\ref{fig:lr_sweep}) with dimensionality fixed to 20, and selection based on loss after 200 epochs.
We selected a learning rate of 0.5 for Euclidean and 5.0 for both \poincare and Lorentz embeddings.
In our initial experiments, we found that other hyperparameters, such as number of negative samples, also affected embeddings performance, but were less influential than learning rate.
See Appendix \ref{app:run_poincare_code} for details on the exact version of the codebase used in our experiments and how it was called.  See the original study~\citep{nickel2017poincare} for additional details on model training and evaluation.

\vspace{.1in}
\noindent {\bf Evaluation.}
Embeddings are evaluated under the original reconstruction error setting.  For each hypernym pair $<u, v>$ in the tree, rank all non-hypernyms along with $v$ by distance from $u$ in the embedded space. The fidelity to which a set of embeddings represents the tree is evaluated according to mean average precision (MAP) and mean rank (MR) of the positive example, $v$, averaged across rankings.

We focus on reconstruction error experiments as they are meant to highlight the capacity of each embedding space.
The ability to generalize out of sample is an orthogonal question, however which we don't address in this work.  This is mostly since the source code to reproduce these results from~\citet{nickel2017poincare} has not yet been released, and the particular folds of heldout edges are also not provided.  As such, we leave reproduction of the link prediction evaluation of Euclidean vs. hyperbolic embeddings to future work.\footnote{At the time of writing, there is an open GitHub issue to provide more details on the out of sample, link prediction evaluation \url{https://github.com/facebookresearch/poincare-embeddings/issues/10}.}

\section{Results}
\label{sec:results}

\begin{table}[t]
  \begin{center}
  \begin{tabularx}{\linewidth}{rrrrrrr}
    \toprule
    {\em dims} & \multicolumn{1}{c}{5} & \multicolumn{1}{c}{10} & \multicolumn{1}{c}{20} & \multicolumn{1}{c}{50} & \multicolumn{1}{c}{100} & \multicolumn{1}{c}{200}  \\
    \midrule
    \multicolumn{7}{l}{\em Mean Average Precision \% (higher is better):}\\
    \multicolumn{7}{c}{Euclidean}\\
     {\em N\&K} & 2.4 & 5.9 & 8.7  & 14   & 16.2 & 16.8 \\
     {ours} &     2.7 & 4.5 & 11.2 & 88.9 & 91.7 & 92.2 \\
    \midrule
    \multicolumn{7}{c}{\poincare}\\
     {\em N\&K} & 82.8 & 86.5 & 85.5 & 86   & 85.7 & 87 \\
     {ours}     & 85.6 & 88.7 & 89.1 & 89.3 & 89.2 & 89.3 \\
    \midrule
    \multicolumn{7}{c}{Lorentz}\\
     {\em N\&K} & 92.3 & 92.8 & \textemdash  & \textemdash & \textemdash & \textemdash \\
     {ours}     & 87.4 & 88.6 & 89.5 & 89.3 & 89.4 & 89.4 \\
    
    \midrule
    \midrule
    \multicolumn{7}{l}{\em Mean Rank (lower is better):}\\
    \multicolumn{7}{c}{Euclidean}\\
     {\em N\&K} & 3542 & 2286 & 1685  & 1281   & 1187 & 1157 \\
     {ours} &     3646 & 1455 & 244 & 1.8 & 1.5 & 1.5 \\
    \midrule
    \multicolumn{7}{c}{\poincare}\\
     {\em N\&K} & 4.9 & 4.0 & 3.8 & 3.9   & 3.9 & 3.8 \\
     {ours}     & 6.8 & 5.6 & 5.2 & 4.9 & 4.9 & 4.9 \\
    \midrule
    \multicolumn{7}{c}{Lorentz}\\
     {\em N\&K} & 3.1 & 2.9 & \textemdash  & \textemdash & \textemdash & \textemdash \\
     {ours}     & 6.6 & 5.5 & 5 & 4.9 & 4.8 & 4.8 \\
  \bottomrule
  \end{tabularx}
  \end{center}
  \caption{Mean average precision and mean rank for reconstructing the {\em WordNet-Nouns} hypernymy graph. {\em N\&K} refers to best published results from~\citet{nickel2017poincare,nickel2018learning}.}
  \label{tab:results_dim_sweep}
\end{table}

Table~\ref{tab:results_dim_sweep} shows the MAP and MR results for the {\em WordNet-Nouns} hierarchy reconstruction task.
There are clear differences between the reproduced and reported performance for Euclidean embeddings.
In the 50 dimensions setting, the reproduced Euclidean embeddings achieve a MAP score of 88.9 compared to 14 in the original study, and an MR score of 1.8 compared to 1,281.
In fact, this MR score for Euclidean embeddings at 50 dimensions is better than that achieved by \poincare and Lorentz embeddings even with 200 dimensions.
With greater than 50 dimensions, Euclidean embeddings outperform both Lorentz and \poincare embeddings according to MR and MAP.


In contrast, the performance of hyperbolic embeddings remain stable across dimensionality and are similar between reproduced and reported results.
The improved reconstruction error of Lorentz embeddings in prior work is likely due to a more comprehensive hyperparameter search.

We also found that performance is robust to random seed for all methods, with a standard deviation of less than a point for both MAP and MR score when training 50-dimensional Euclidean, \poincare, or Lorentz embeddings (\Cref{tab:app:dim50_random_restarts}).



\begin{table}[t]
\begin{center}
\begin{tabular}{lrr}
\toprule
  Embedding & MAP & MR \\
\midrule
  Euclidean & 89.2 $\pm$ 0.33 & 1.8 $\pm 0.00$ \\  
  \poincare & 89.4 $\pm$ 0.09 & 4.9 $\pm$ 0.01 \\
  Lorentz   & 89.4 $\pm$ 0.26 & 4.9 $\pm$ 0.10 \\ 
  \bottomrule
\end{tabular}
    \caption{Mean and standard deviation of MAP and MR for 50-dimensional embeddings across three different random restarts.}
    \label{tab:app:dim50_random_restarts}
\end{center}
\end{table}


\section{Analysis}
\label{sec:analysis}

\begin{table*}[t!]
  \begin{center}
  \begin{tabularx}{0.6\linewidth}{lrrrrrr}
    \toprule
    {\em dims} & \multicolumn{1}{c}{5} & \multicolumn{1}{c}{10} & \multicolumn{1}{c}{20} & \multicolumn{1}{c}{50} & \multicolumn{1}{c}{100} & \multicolumn{1}{c}{200}  \\
    \midrule
    \multicolumn{7}{c}{\em Mean Average Precision \% (higher is better):}\\
     {\em N\&K} & 2.4 & 5.9 & 8.7  & 14.0   & 16.2 & 16.8 \\
     {ours: unit norm} & 2.4 & 5.0    & 7.6 & 10.6 & 12.0 & 12.5 \\
       \midrule
     {ours: no norm}   & 2.7 & 4.5 & 11.2 & 88.9 & 91.7 & 92.2 \\
    \midrule
    \multicolumn{7}{c}{\em Mean Rank (lower is better):}\\
     {\em N\&K} & 3542 & 2286 & 1685  & 1281   & 1187 & 1157 \\
     {ours: unit norm} & 3807 & 2275 & 1697 & 1276 & 1184 & 1159 \\
       \midrule
     {ours: no norm}   & 3646 & 1455 & 244 & 1.8 & 1.5 & 1.5 \\
  \bottomrule
  \end{tabularx}
  \end{center}
  \caption{MAP and MR for reconstructing the {\em WordNet-Nouns} hypernymy graph. 
  Results from {\em N\&K}~\citep{nickel2017poincare,nickel2018learning} and our reproduction with Euclidean embeddings constrained to unit norm are similar.}
  \label{tab:results_dim_sweep_max_norm_1}
\end{table*}

In \Cref{sec:results}, we show it is possible to learn Euclidean embeddings that can reconstruct \emph{WordNet-Nouns} more faithfully than similarly trained hyperbolic embeddings with at least 50 dimensions.
In this section, we discuss possible reasons for the discrepancy between reported and reproduced Euclidean embeddings performance.
We first posed this question to the authors themselves~\citep{nickel2017poincare} who clarified on GitHub that the difference in performance for Euclidean embeddings in their published manuscript was due to a regularization method used at the time.
They further added that in the released code they ``\textit{disabled this regularization by default and it turned out to work better}''.\footnote{Author response regarding difference in Euclidean performance \url{https://github.com/facebookresearch/poincare-embeddings/issues/35\#issuecomment-685261354}}
Follow-up questions regarding the details of this regularization method have yet to be addressed, at the time of writing.\footnote{Question seeking further information regarding regularization \url{https://github.com/facebookresearch/poincare-embeddings/issues/35\#issuecomment-685261354} and a relevant comment in \tt{issuecomment-735209781}.}

\paragraph{Constraining Euclidean Embedding Norm} We speculate that the authors may have normalized the Euclidean embeddings to constrain them to lie within a unit 2-norm ball, similar to how \poincare embeddings are trained~\citep[Section~3.1]{nickel2017poincare}.  Note that while projection into the unit ball is necessary to learn valid \poincare embeddings, Euclidean embeddings require no such constraint.

The released source code actually supports projecting Euclidean embeddings into the unit ball after each iteration, but this is disabled by default, with the argument \emph{max\_norm} set to \emph{None}.\footnote{L2 normalization disabled by default: \url{https://github.com/facebookresearch/poincare-embeddings/blob/4c7316b/hype/manifolds/euclidean.py\#L16}.}
To test whether this constraint has an effect on embedding quality, we train Euclidean embeddings constrained to the unit ball by setting \emph{max\_norm} to 1. \Cref{tab:results_dim_sweep_max_norm_1} shows that reconstruction scores for Euclidean embeddings constrained to the unit ball are much closer to those published in~\citet{nickel2017poincare,nickel2018learning} than unconstrained Euclidean embeddings.

\paragraph{Varying Norm Constraints} To explore the impact of norm constraints, 
 we conduct 
further 
experiments on training 100 dimensional Euclidean embeddings. We vary the \emph{max\_norm} setting between 1 and 10, allowing the Euclidean embeddings to grow larger during training.
As an alternative method for controlling the norm of the embeddings, we also vary the strength of an L2 penalty as an additional term in the loss. 
MAP and MR scores improve as the \emph{max\_norm} setting is increased, the norm constraint is relaxed (\Cref{tab:app:dim100_euclidean_regularization}).  In fact, setting \emph{max\_norm} to 5 yields Euclidean embeddings that achieve similar reconstruction performance as unconstrained embeddings.

We found that including an L2 regularization penalty in the loss has little effect on the final reconstruction scores. Thus, although we suspect unnecessary renormalization of the Euclidean embeddings may have caused the poor reconstruction performance reported in prior work, we defer to the authors of the original study for confirmation.


\begin{table}[ht]
\begin{center}
\begin{tabular}{lrr}
\toprule
  Constraint & MAP & MR \\
\midrule
  None (default) & 91.7 & 1.5 \\
  L2 regularization = 0.01 & 92.3 & 1.5 \\ 
  L2 regularization = 1.0 & 92.3 & 1.5 \\ 
  L2 regularization = 100.0 & 92.2 & 1.5 \\ 
  max norm = 1 & 11.2 & 1150.0 \\
  max norm = 2 & 38.3 & 40.3 \\
  max norm = 5 & 92.3 & 1.5 \\
  max norm = 10 & 92.3 & 1.5 \\
  \midrule
  {\em N\&K}     & 16.2   & 1187.3 \\
  \bottomrule
\end{tabular}
    \caption{MAP and MR for 100 dimensional embeddings. 
    We include results with no regularization or constraints (row 1) and with L2 regularization of varying degree.
    \textit{max norm = k} means that embeddings are projected back into a radius $k$ 2-norm ball after each iteration.
    {\em N\&K} refers to best published results from~\citet{nickel2017poincare,nickel2018learning}.
    }
    \label{tab:app:dim100_euclidean_regularization}
\end{center}
\end{table}

\section{Conclusion}
\label{sec:conclusion}
In our reproduction of the experiments in Table 1 of \citet{nickel2017poincare}, we find that Euclidean actually outperform \poincare embeddings when allowed a moderate number of dimensions.
This is a realistic number of dimensions for typical non-contextual word embeddings, and a far lower dimensionality than subword token embeddings used in pretrained transformer language models. 
For example, released GloVe embeddings range from 50 to 300 dimensions \citep{pennington2014glove} and BERT base uses 768-dimensional subword embeddings \citep{devlin2019bert}.

Nevertheless, the strong performance of hyperbolic embeddings in very low dimensions (less than 20) highlights their main strength: succinctly embedding nodes in hierarchically structured graphs with tight limitations on embedding size. 

However, part of~\citet{nickel2017poincare}'s impact came from the astounding gains from hyperbolic embeddings over Euclidean across a wide range of embedding widths. 
Subsequent application of \poincare embeddings often report mixed results when using non-Euclidean vs. Euclidean embeddings in downstream tasks \citep{dhingra-etal-2018-embedding,lopez-etal-2019-fine}. 
We hope that this reproduction will serve as a valuable reference for others who are just beginning to explore hyperbolic embeddings.



\section*{Acknowledgments}
\label{sec:thanks}
We thank Maximilian Nickel and Douwe Kiela for their helpful feedback while carrying out this study and for making the source code available.
We also thank the workshop organizers for providing an avenue for such work to be published and the anonymous reviewers whose feedback helped us improve this work.
Thanks also to the GitHub users who reported similar issues in the source code repository, which motivated us to publish this study.

\bibliography{main}
\bibliographystyle{acl_natbib}

\newpage
\clearpage

\appendix 

\section{Call to \poincare Embedding Trainer}
\label{app:run_poincare_code}

Sample call to train embeddings and run reconstruction evaluation:

\begin{lstlisting}[language=bash]
### Parameters
LR=0.5
DIM=20
MANIFOLD="euclidean"

# Train model
python -m hype.embed \
  -checkpoint model.bin
  -dset wordnet/noun_closure.csv
  -epochs 1500
  -negs 50
  -burnin 20
  -dampening 0.75
  -ndproc 4
  -eval_each 100 \
  -fresh
  -sparse
  -burnin_multiplier 0.01
  -neg_multiplier 0.1
  -lr_type constant
  -train_threads 1
  -dampening 1.0
  -batchsize 50
  -manifold ${MANIFOLD}
  -dim ${DIM}
  -lr ${LR}

python reconstruction.py model.bin.1499
\end{lstlisting}

We use the \texttt{poincare-embeddings} implementation~\url{https://github.com/ facebookresearch/poincare-embeddings}
at commit 
  \texttt{4c7316b14dce3b89e6a2d0c7994d418dffb42c94}.

\end{document}